\documentclass[preprint,12pt]{elsarticle}

\usepackage{amssymb}
\usepackage{amsmath}
\usepackage{comment}
\usepackage{amsfonts}
\usepackage{amssymb}
\usepackage{graphicx}
\usepackage{mathrsfs}
\usepackage[english]{babel}
\usepackage{combelow}
\usepackage{booktabs} 
\usepackage{tabularx} 
\newcommand{\tableheadline}[1]{\multicolumn{1}{l}{\textbf{#1}}}

\usepackage[latin1]{inputenc}

\newcommand{\dGLI}{\text{dGLI}}

\usepackage{color}

\journal{Engineering Applications of Artificial Intelligence. }

\begin{document}

\begin{frontmatter}



\title{The dGLI Cloth Coordinates:\\A Topological Representation for\\Semantic Classification of Cloth States}


\author[mymainaddress]{Franco Coltraro\corref{mycorrespondingauthor}}
\cortext[mycorrespondingauthor]{Corresponding author}
\ead{franco.coltraro@upc.edu}
\author[mysecondaryaddress]{Josep Fontana}

\author[mysecondaryaddress]{Jaume Amor\'os}
\author[mymainaddress,mysecondaryaddress]{Maria Alberich-Carrami\~nana}
\author[mymainaddress]{J\'ulia Borr\`as }
\author[mymainaddress]{Carme Torras}

\address[mymainaddress]{Institut de Rob\`otica i Inform\`atica Industrial, CSIC-UPC, \\
C/ Llorens i Artigas 4-6, 08028, Barcelona, Spain.}
\address[mysecondaryaddress]{Departament de Matem\`atiques, Universitat Polit\`ecnica de Catalunya, Barcelona, Spain.}

\begin{abstract}


Robotic manipulation of cloth is a highly complex task because of its infinite-dimensional shape-state space that makes cloth state estimation very difficult. In this paper we introduce the dGLI \textit{Cloth Coordinates}, a low-dimensional representation of the state of a rectangular piece of cloth that allows to efficiently distinguish key topological changes in a folding sequence, opening the door to efficient learning methods for cloth manipulation planning and control. Our representation is based on a directional derivative of the \textit{Gauss Linking Integral} and allows us to represent both planar and spatial configurations in a consistent unified way. The proposed dGLI Cloth Coordinates are shown to be more accurate in the classification of cloth states and significantly more sensitive to changes in grasping affordances than other classic shape distance methods. Finally, we apply our representation to real images of a cloth, showing we can identify the different states using a simple distance-based classifier.
\end{abstract}




\end{frontmatter}


\section{Introduction}


Textile objects are important and omnipresent in many relevant scenarios of our daily lives, like domestic, healthcare or industrial contexts. However, as opposed to rigid objects, whose pose is fixed with position and orientation, textile objects are challenging  to handle for robots because they change shape under contact and motion, resulting in an infinite-dimensional configuration space. This huge dimensional jump makes existing perception and manipulation methods difficult to apply to textiles. Recent reviews on cloth manipulation like \cite{sanchez2018robotic, yin2021modeling} agree on the need to find a simplified representation that enables more powerful learning methods to solve different problems related to cloth manipulation. 

\begin{figure}[b!]
	\centering
	\includegraphics[width=\linewidth]{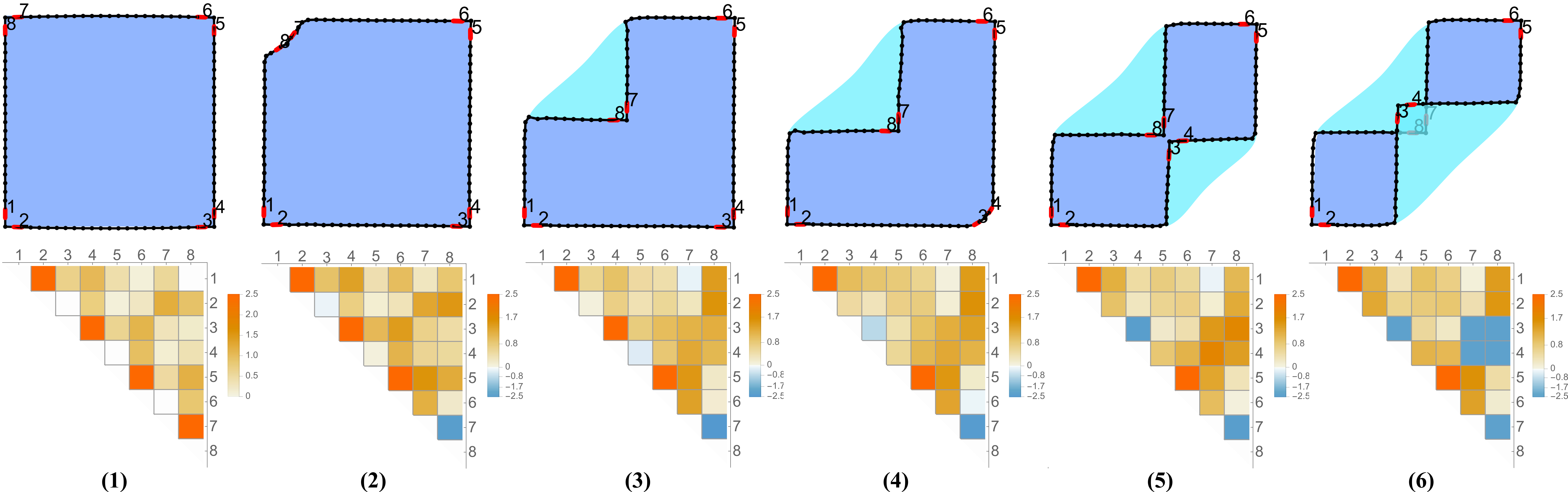}
	\caption{Folding sequence of a quadrangular cloth with its associated dGLI \textit{cloth coordinates}, represented as upper triangular matrices. Each matrix element $m_{ij}$ is a geometrical value corresponding to the $dGLI$ between the segments $i$ and $j$ highlighted in red in the corresponding folded state of the cloth. Notice how some values of the matrix change sign when corners are folded or cross each other.  }
	\label{fig:framesWithMatrix}
\end{figure}

Different representations have been used in the literature of cloth manipulation, e.g. silhouette representations \cite{miller2012geometric} or contours \cite{Doumanoglou2016}, assuming the high-level reasoning on cloth states was given. More modern end-to-end learning approaches use RGB-D images as direct input  \cite{matas2018sim, seita2020deep,jangir2020dynamic,lippi2020latent,tanaka2018emd}, but only very simple actions can be defined due to the limited state representation. In addition, these methods need large amounts of real or simulated data that are expensive to obtain and label, as no underlying previous knowledge is used to understand the geometric relationship between different states.

Therefore, finding a low-dimensional representation for cloth based on low-level features remains an active open problem, while the high-level aspect of understanding cloth deformation is still almost unexplored. On the other hand, to enable reasoning, abstraction and planning, rigid object manipulation applies object recognition methods in order to link objects to actions/affordances \cite{worgotter2013simple,BOUSQUETJETTE201733}. Contacts are estimated among the objects to recognize states such as ``on top of", ``inside of" \cite{aksoy2011learning}. However, when it comes to cloth manipulation, no work has explored the semantic state identification that could lead to particular actions depending on the task in mind. For simpler deformable objects like a box with an articulated lid, the open configuration clearly allows the action of closing the box or picking something from inside. An equivalent example for cloth would be to recognize a folded corner that needs to be either flattened back if the task is to lay it flat on the table, or pick it up if the task is folding. In this context, we wish to classify the configuration space of a piece of cloth in macro-states (or just states), where each state is the set of cloth configurations that can be manipulated in the same way,
i.e., that have similar grasping affordances.

In this work, we present a coordinate representation of the configuration of a rectangular cloth as an upper triangular matrix form (see Figure \ref{fig:framesWithMatrix}).  This representation can be computed with a closed-form formula from low-level features of the cloth, mainly the position of its border, and enables the recognition and classification of high-level states, since we can define a distance between cloth configurations. That allows us to classify different configurations into states that we identify as ``different", meaning that they afford different actions.

Our results show we can identify the topologically relevant changes along folding sequences, just using a distance between our representations as vectors. This same distance allows us to distinguish 12 different classes generated with a cloth simulator. These are promising results towards a low-dimensional representation that can be used for high-level identification of states, but is still linked to low-level features such as the location of the border, which are fundamental to execute physical actions. 


One limitation of our approach is that it assumes that the cloth boundaries are known. However, perception algorithms start to show solutions to overcome this problem. For instance, in \cite{RAMISA2014246} a method is developed to detect parts of clothes suitable for grasping. More recently, the deep-learning approach presented in \cite{qian2020cloth} can identify corners and edges, but does not yet identify the full border. Our group is working on deep-learning methods to \textit{hallucinate} the full border given an image of the cloth. Despite this limitation, our approach as it stands can be fully used in simulation, for instance to automatically label cloth states.

Our coordinates are based on a topological index, the \textit{Gauss Linking Integral} ($GLI$). This index has been used in the past for robotic manipulation \cite{ivan2013topology, zarubin2012hierarchical,pokorny2013grasping, stork2013integrated, stork2013topology} but can only be applied to 3D curves. 
For a pair of almost coplanar curves, as the boundary curves of a folded garment, the $GLI$ vanishes and it ceases to be informative.
In order to consistently consider 2D curves as well as 3D curves, we introduce in this work the concept of the directional derivative of the $GLI$, $dGLI$, applied to a pair of curves. The $dGLI$ is symmetric on the curves and it only depends on the relative position between them.
We assign the $\dGLI$  \emph{Cloth Coordinates} to a state of a garment as follows: first select a subset of edges (it may contain the whole of them) from a discretization of the boundary of the garment; then fix an ordering on these edges and compute the $dGLI$ between any pair of edges in their spatial position of the current state of the garment; this gives a symmetric matrix from which only the upper triangular part is taken in order to avoid redundancies; the $\dGLI$ cloth coordinates of the state are precisely the entries of this upper triangular matrix (see Figure \ref{fig:framesWithMatrix}).
Our resulting representation can be computed efficiently and it is invariant under isometric movements of the garment (i.e. rotations and translations) leaving invariant a distinguished direction which is normal to a predominant plane in the scene (as for instance a table used as support for the manipulations).

This article is structured as follows: in the next section we present preliminary concepts used in the paper, such as the Gauss Linking Integral, and we explain its limitations in a planar setting. 
Then, in Section \ref{sec:theory} we introduce the novel concept of the directional derivative of the $GLI$ which is also applicable in a flat space. We derive first an expression for the $GLI$ of two segments, then we prove that we can perturb the segments slightly to obtain information when they are co-planar and we explain how to apply this to a full meshed cloth. Next we study some of the properties of this new index by applying it to a data-base of cloth configurations taken from simulated folding sequences, to then experimentally test the index on real images. Finally, we discuss the obtained results and draw some conclusions in the last section. 

\section{Preliminaries and related work} \label{sec:preliminaries}


Given two non-intersecting 3D-space curves $\gamma_{1}$, $\gamma_{2}$ parameterized by $x(s)$ and $y(t)$, respectively, with $s,t\in I=[0,1]$, the \emph{Gauss Linking Integral} between them, $GLI$ for short, is
\begin{equation*}
   GLI(\gamma_{1},\gamma_{2})
 =  \frac{1}{4\pi}\int_I\int_I \frac{(y(t)-x(s))\cdot(y'(t)\times x'(s))}{|y(t)-x(s)|^3}dtds 
\end{equation*}
or written in a compact way
\begin{equation} \label{eq:GLIDef}
   GLI(\gamma_{1},\gamma_{2})
 =  \frac{1}{4\pi}\int\int\frac{(\gamma_{2}-\gamma_{1})\cdot [\gamma_{2}' \times \gamma_{1}']}{\Vert \gamma_{2}-\gamma_{1} \Vert^3} \, .
\end{equation}
This double integral is invariant under re-parameterizations of the curves. In the case that both curves  $\gamma_{1}$ and $\gamma_{2}$ are closed and smooth, their $GLI$ is integer valued (due to the chosen normalization factor $\frac{1}{4\pi}$) and it is an invariant of the topology of the embedded curves (see \cite{Aldinger-Klapper-Tabor1995}). 

Historically, the $GLI$ was first introduced by Gauss, presumably related to his works on magnetism (according to \cite{Ricca-Nipoti2011}) or on astronomy (according to \cite{Epple1998}).
Considering the $GLI (\gamma ,\gamma)$ of twice the same non-self-intersecting smooth curve $\gamma$, then the double integral (taking the domain of integration outside the diagonal of $I \times I$) defines another geometric invariant of the curve, known as \emph{writhe} or \emph{writhing number} of $\gamma$.
Despite their resemblance, the $GLI$ and the writhe measure different quantities: consider a normal vector field $v$ of length $\epsilon >0$ on $\gamma$, and the curve $\gamma _{v}$ of endpoints of the vector field $v$, which is embedded and in one-to-one smooth correspondence with $\gamma$ for sufficiently small $\epsilon$. Then the $GLI$ of these two close copies of the same $\gamma$ differs from the  writhe in $GLI (\gamma ,\gamma_{v}) - GLI (\gamma ,\gamma) $ equal to the total twist of $v$. This result is known as the C\u{a}lug\u{a}reanu-White-Fuller theorem (see \cite{Pohl1980}). However, both indexes, $GLI$ and writhe, are non-informative for planar curves, since they both vanish.

The $GLI$ has been used for many applications after a version of the above formula for polygonal curves appeared in the context of DNA protein structures \cite{levitt1983protein}, with additional efficient formulations given in  \cite{KleninandLangowski2000} from which we have chosen the following: given a discretization of the curves into $N$ and $M$ segments, that is, $\gamma_1=\{ \gamma_{P_iP_{i+1}}, i=1,\dots, N\}$ and $\gamma_2=\{ \gamma_{Q_iQ_{i+1}}, i=1,\dots, M\}$,
where each segment is parameterized as $\gamma_{AB}(s)=A+s\Vec{AB}$ for $s\in [0,1]$, then the $GLI$ between both curves is 
\begin{equation}\label{eq:levitt}
    GLI(\gamma_1,\gamma_2) =\frac{1}{4\pi}\sum_{i=1}^N\sum_{j=1}^M GLI(\gamma_{P_iP_{i+1}},\gamma_{Q_iQ_{i+1}}) 
\end{equation}
where the $GLI$ between a pair of segments $\gamma_{AB}$ and $\gamma_{CD}$ is computed as
\begin{equation}\label{eq:GLI2Segments}
\begin{split}
 GLI(\gamma_{AB},\gamma_{CD})=& \arcsin(\Vec{n}_A\Vec{n}_D) +  \arcsin(\Vec{n}_D \Vec{n}_B)  \\
                     +& \arcsin(\Vec{n}_B\Vec{n}_C) +\arcsin(\Vec{n}_C \Vec{n}_A) 
 \end{split}
 \end{equation}
with
\begin{equation*}
\begin{split}
        \Vec{n}_A=\| \Vec{AC} \times \Vec{AD} \|\text{,      }  &  \Vec{n}_B=\| \Vec{BD} \times \Vec{BC} \|, \\
        \Vec{n}_C=\| \Vec{BC} \times \Vec{AC} \|\text{, and  }&\Vec{n}_D=\| \Vec{AD} \times \Vec{BD} \|.
\end{split}
\end{equation*}

The above discrete formula  was used by Ho \cite{ho2011thesis} to identify and synthesize animated characters in intertwined positions \cite{ho2010controlling, ho2011thesis}. In the context of robotics, the $GLI$ has been applied to representative curves of the workspace to guide path planning through holes \cite{ivan2013topology, zarubin2012hierarchical}, for guiding caging grasps in \cite{pokorny2013grasping, stork2013integrated, stork2013topology}, and for planning humanoid robots motions, using the $GLI$ to guide reinforcement learning \cite{yuan2019reinforcement}. In this work, for the first time, we develop a further analysis of the notion to be able to apply it to planar or almost planar curves, which opens the door to a wider spectrum of applications.





\section{Derivation of the Cloth Coordinates} \label{sec:theory}
As we have mentioned above, the $GLI$ of two coplanar curves vanishes; so for many configurations of robotic interest --- configurations where the cloth is  nearly flat on a table, ready to be folded or already folded---  the $GLI$ does not provide much information. Our aim in this section is therefore to develop a similar index which is able to distinguish planar configurations. We shall see that a natural index to consider is in fact a directional derivative of the $GLI$, but to arrive at such an index we must first make a few observations about the $GLI$ when applied to pairs of segments as in Eq. (\ref{eq:GLI2Segments}); since the class of curves we will be working with computationally are piece-wise linear.

\subsection{$GLI$ of two segments}
Since two segments $AB$ and $CD$ are uniquely defined by the four endpoints $A,B,C,D\in\mathbb{R}^3$, the $GLI$ of two segments computed in Eq. (\ref{eq:GLI2Segments}) can be viewed as a function from $(\mathbb{R}^3)^4 \equiv \mathbb{R}^{12}$ to $\mathbb{R}$. To emphasize that from now on we are considering segments we define $\mathcal{G}:\mathbb{R}^{12}\rightarrow \mathbb{R}$ as 
$$
\mathcal{G}(A,B,C,D)=GLI(\gamma_{AB},\gamma_{CD}).
$$
Note that technically $\mathcal{G}$ is not defined in the whole of $\mathbb{R}^{12}$, since it is not defined when $\gamma_{AB}$ and $\gamma_{CD}$ intersect. 
Next we will find a reformulation for $\mathcal{G}$ wherever it is defined. 

Notice that the numerator in the integral expression of the $GLI$ of Eq. (\ref{eq:GLIDef}) is  constant (for any $t$ and $s$) and equals
\begin{align*}
    (\gamma_{CD}-\gamma_{AB})\cdot &[\gamma_{CD}' \times \gamma_{AB}'] = \\ 
    &= (\Vec{AC} + t\Vec{CD} - s\Vec{AB}) \cdot [\Vec{CD}\times\Vec{AB}] = \\
    &= \Vec{AC} \cdot [\Vec{CD}\times\Vec{AB}]= \\
    &= \Vec{AC} \cdot [(\Vec{CA}+\Vec{AD})\times\Vec{AB}] =\\
    &  = \Vec{AC} \cdot [\Vec{AD}\times\Vec{AB}] = \Vec{AB} \cdot [\Vec{AC}\times\Vec{AD}]=\\
    & = \det(\Vec{AB},\Vec{AC},\Vec{AD})
\end{align*}
the signed volume of the tetrahedron $ABCD$  multiplied by $6$. By writing 
\begin{equation*}
    \mathcal{V}(A,B,C,D)=
    \det(\Vec{AB},\Vec{AC},\Vec{AD})
\end{equation*}
and \begin{equation*}
    \mathcal{I}(A,B,C,D)=\frac{1}{4\pi}\int\int\frac{1}{\Vert \gamma_{CD}-\gamma_{AB} \Vert^3} \, ,
\end{equation*}
we have
\begin{equation}\label{segmentG}
    \mathcal{G} = \mathcal{V} \cdot \mathcal{I} \, .
\end{equation}


Now, it is clear that the $GLI$ vanishes when the segments are coplanar because $\mathcal{V}=0$.
Being $\mathcal{G}$ the product of two differentiable functions and hence differentiable, it makes sense to consider its directional derivative.


\subsection{Directional derivative of $\mathcal{G}$}\label{subsec:dGLI}

Let $v_A,v_B,v_C,v_D\in \mathbb{R}^3$ be arbitrary directions in which we will perturb the corresponding vertices of the pair of segments. Then $v = (v_A,v_B,v_C,v_D)\in \mathbb{R}^{12}$ defines a direction in the ``aggregate segment space". 
The directional derivative $ \partial_v \mathcal{G}(A,B,C,D)$ of $\mathcal{G}$ at the point $(A,B,C,D)$ in the direction of $v$ is
\begin{equation*}
    \lim_{\varepsilon \rightarrow 0}\frac{\mathcal{G}((A,B,C,D) + \varepsilon (v_A,v_B,v_C,v_D)) - \mathcal{G}(A,B,C,D)}{\varepsilon} ,
\end{equation*}
which can be equivalently written as
\begin{equation*} 
\lim_{\varepsilon \rightarrow 0}\frac{GLI(\gamma_{A^*
B^*} ,\gamma_{C^*
D^*} ) -  GLI(\gamma_{A
B} ,\gamma_{C
D})}{\varepsilon} ,
\end{equation*}
where $A^*= A+ \varepsilon v_A$, $B^*= B+ \varepsilon v_B$, $C^*= C+ \varepsilon v_C$ and $D^*= D+ \varepsilon v_D$. The $GLI$ function in the numerator of $\partial_v \mathcal{G}$ may be effectively computed using Eq. (\ref{eq:GLI2Segments}). 

From Eq. (\ref{segmentG}) and by the product rule

\begin{equation*}
    \partial_v \mathcal{G} = \partial_v (\mathcal{V})\mathcal{I} + \mathcal{V}\partial_v (\mathcal{I}),
\end{equation*}
hence $ \partial_v \mathcal{G} = \partial_v (\mathcal{V})\mathcal{I}$ when the segments $\gamma_{AB}$ and $\gamma_{CD}$ are coplanar. 

This new index $\partial_v \mathcal{G}$, being informative for both coplanar and non-coplanar pairs of segments, will be useful if the right choice of $v$ is made, as we will discuss next.

By definition $\partial_v\mathcal{G}$ is invariant under translations, rotations and scalings if $v$ is rotated and scaled accordingly. These properties are a consequence of the fact that the $GLI$ is invariant under such transformations. However, for a fixed choice of $v$, $\partial_v\mathcal{G}$ will not be invariant under rotations or scalings in general. For instance, no fixed choice of $v$ can make $\partial_v\mathcal{G}$ invariant under scalings, since scaling by a factor of $\lambda$ scales $\mathcal{I}$ by $\frac{1}{\lambda^3}$ and $\nabla\mathcal{V}$ by $\lambda^2$, and similarly scales $\nabla\mathcal{I}$ by $\frac{1}{\lambda^4}$ and $\mathcal{V}$ by $\lambda^3$, resulting in scaling $\nabla\mathcal{G}$ by $\frac{1}{\lambda}$. Depending on what this index is used for one must keep this scaling relationship in mind or alternatively choose $v$ depending on the segments. However, the distance we will use to compare different cloth states only depends on the correlation of values of the coordinates more than on the magnitude. That is why we can ignore the scaling factor that would appear when comparing two garments of different sizes (e.g. because of different meshings).

The choice of $v$ is highly task-specific. Given the nature of our task --- classifying planar cloth configurations based on affordances --- it is natural to perturb the vertices in the direction normal to the table plane. Making such a choice of $v$ does in fact make $\partial_v\mathcal{G}$ invariant under rotations and translations of the $XY$ plane, which is desirable for our purposes since such movements of a cloth configuration have the same affordances. Furthermore, to conserve the symmetry $\partial_v\mathcal{G}(A,B,C,D) = \partial_v\mathcal{G}(C,D,A,B)$, we must perturb $A$ and $C$ by the same amount and direction, and the same is the case with $B$ and $D$. Finally, it is easy to see that in fact perturbing $A$ and $C$ both by the same amount normal to the plane yields the same result as perturbing $B$ and $D$ by the same amount in the opposite direction, so it really only makes sense to perturb $A$ and $C$, or $B$ and $D$, but not both pairs, and doing one or the other is equivalent except for a sign change.

In summary, the most natural choice of $v$ in our case is 
\begin{equation}\label{eq:chosenV}
   v:=(\textbf{0},e_3,\textbf{0},e_3) 
\end{equation}
(or $v=(e_3,\textbf{0},e_3,\textbf{0})$, which is equivalent except for a sign change) where $e_3=(0,0,1)$ is the normal to the plane of the table on which the cloth lies. 

\subsection{$dGLI$ of two segments}

The $dGLI$ between two non-intersecting segments $\gamma_{AB}$ and $\gamma_{CD}$, is defined as
\begin{equation} \label{eq:dGLI2Segments}
dGLI(\gamma_{AB},\gamma_{CD}) := 
\lim_{\varepsilon \rightarrow 0}\frac{GLI(\gamma_{A
B^*} ,\gamma_{C
D^*} ) -  GLI(\gamma_{A
B} ,\gamma_{C
D})}{\varepsilon} ,
\end{equation}
where $B^*= B+ \varepsilon e_3$, $D^*= D+ \varepsilon e_3$ and $e_3=(0,0,1)$. Each $GLI$ function can be computed using Eq. (\ref{eq:GLI2Segments}). Therefore, we have that
\begin{equation*} 
dGLI  (\gamma_{AB},\gamma_{CD}) = \partial_v \mathcal{G}(A,B,C,D),
\end{equation*}
taking $v$ as in Eq. (\ref{eq:chosenV}).

We have analyzed numerically the limit defined in Eq. (\ref{eq:dGLI2Segments}), and have found that it is sufficiently stable to be computed as 
\begin{equation*}
     dGLI(\gamma_{AB},\gamma_{CD}) \approxeq   \frac{GLI(\gamma_{A
B^*} ,\gamma_{C
D^*} ) -  GLI(\gamma_{A
B} ,\gamma_{C
D})}{\varepsilon}
\end{equation*}
for a chosen $\varepsilon$ sufficiently small, as for instance $\varepsilon\approx 10^{-8}$, which is the value taken in our experiments.


In practical implementations we may well be computing the $dGLI$ between segments $\gamma_{AB}$ and $\gamma_{CD}$ that are very close to intersect, and then  $dGLI(\gamma_{AB},\gamma_{CD})$ goes to infinity. As having such big quantities can dominate values of metrics and distances in a non-representative way, in practice we set a maximum value to the $dGLI$ once it surpasses a fixed threshold. 

Since we are now equipped with a geometric index for pairs of  segments, we are ready to introduce our  cloth coordinates which will parametrize the shape-state space of a piece of cloth.

\subsection{Definition of the $\dGLI$ Cloth Coordinates}

We assign the $\dGLI$  \emph{Cloth Coordinates} to a configuration $\mathcal{C}$ of a garment as follows:
discretize the garment's boundary (in  vertices and edges), select an ordered subset of edges and consider the segment $S_i$ of the spatial position in configuration $\mathcal{C}$ of the $i$-th edge of the discretization, $\mathcal{S}_\mathcal{C}=\{ S_i, i=1,\dots, m\}$. The $\dGLI$  \emph{Cloth Coordinates} of configuration $\mathcal{C}$ is the upper triangular matrix 
\begin{equation} \label{eq:dGLICoord}
    \text{dGLI}(\mathcal{C}) = \big( dGLI(S_i,S_j) \big)_{S_i,S_j \in \mathcal{S}_\mathcal{C}, i>j}.
\end{equation}
To get an intuitive sense of what these upper triangular matrices look like for some cloth configurations, see the examples in Figure \ref{fig:framesWithMatrix}.
If we were interested in a general direction $v$, we would take the $\dGLI _v$ Cloth Coordinates
\begin{equation*}
    \text{dGLI}_{v}(\mathcal{C}) = \big( dGLI_v (S_i,S_j) \big)_{S_i,S_j \in \mathcal{S}_\mathcal{C}, i>j}.
\end{equation*}
Note that the full matrix when taking the whole of edges of the discretization is the equivalent rationale than computing the \textit{GLI} of a polynomial curve used in \cite{ho2011thesis,levitt1983protein}, where the \textit{GLI} of all pairs of segments of the curves where first assembled in what was called the \textit{GLI matrix} \cite{ho2011thesis}.

\begin{figure}[htb]
	\centering
	\includegraphics[width=0.3\linewidth]{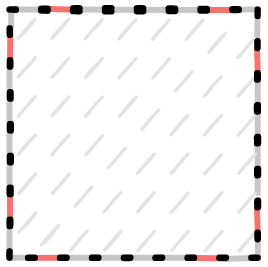}
	\caption{The subset of chosen segments are marked red.}
	\label{fig:selection_edges}
\end{figure}

The subset  of edges chosen in the discretization would depend on the task one wants to carry out; tasks which demand finer distinctions between configurations of a similar class would require a greater subset of segments. For our task of classifying the configurations into relatively broad classes, we found experimentally that a good choice of segments are the eight segments adjacent to the corner segments, marked red in Figure \ref{fig:selection_edges}. This is a small subset that is nevertheless enough to provide an accurate affordance-based classification of near-to-flat configurations.

The upper triangular matrix in Eq. (\ref{eq:dGLICoord}), sorted as a vector, is a coordinate system that reduces the high dimensionality of the configuration space of cloth states to a mere $\frac{m(m-1)}{2}$ dimensional space. In our case $m = 8$, so this comes out to 28 dimensions. 
This reduction in dimensionality is well-suited and informative enough for practical purposes, as the validation results in next section will show.

\section{Results} \label{sec: results}

\begin{figure}[h!]
	\centering
	\includegraphics[width=0.9\linewidth]{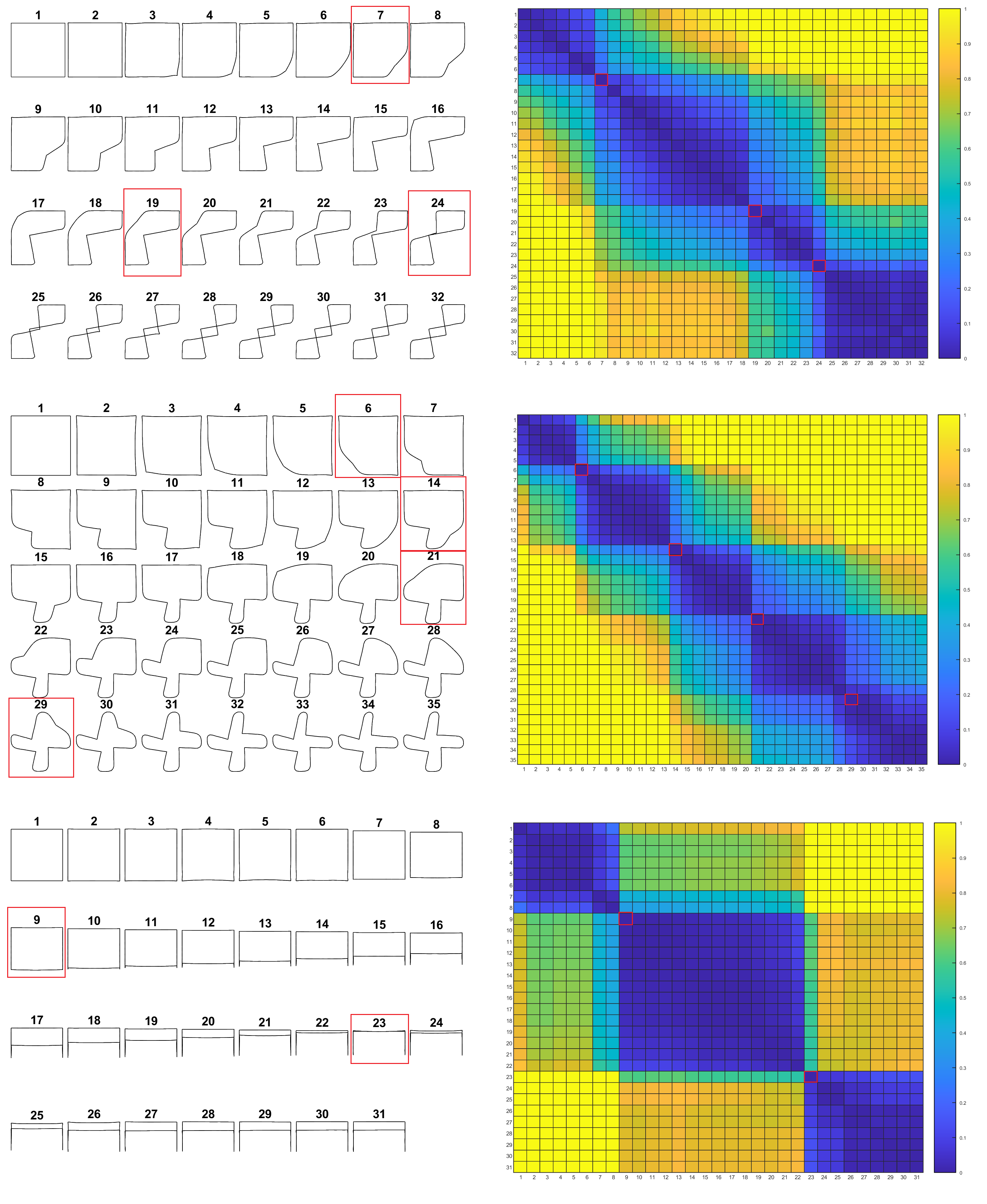}
	\caption{Study of the index during 3 folding sequences. In the left column we show a representation of the cloth frames, and in the right column the confusion matrix of all of them. In red we highlight the clear class changes that can be identified.}
    \label{fig:sequences}
\end{figure}
\begin{figure}[h!]
	\centering
	\includegraphics[width=\linewidth]{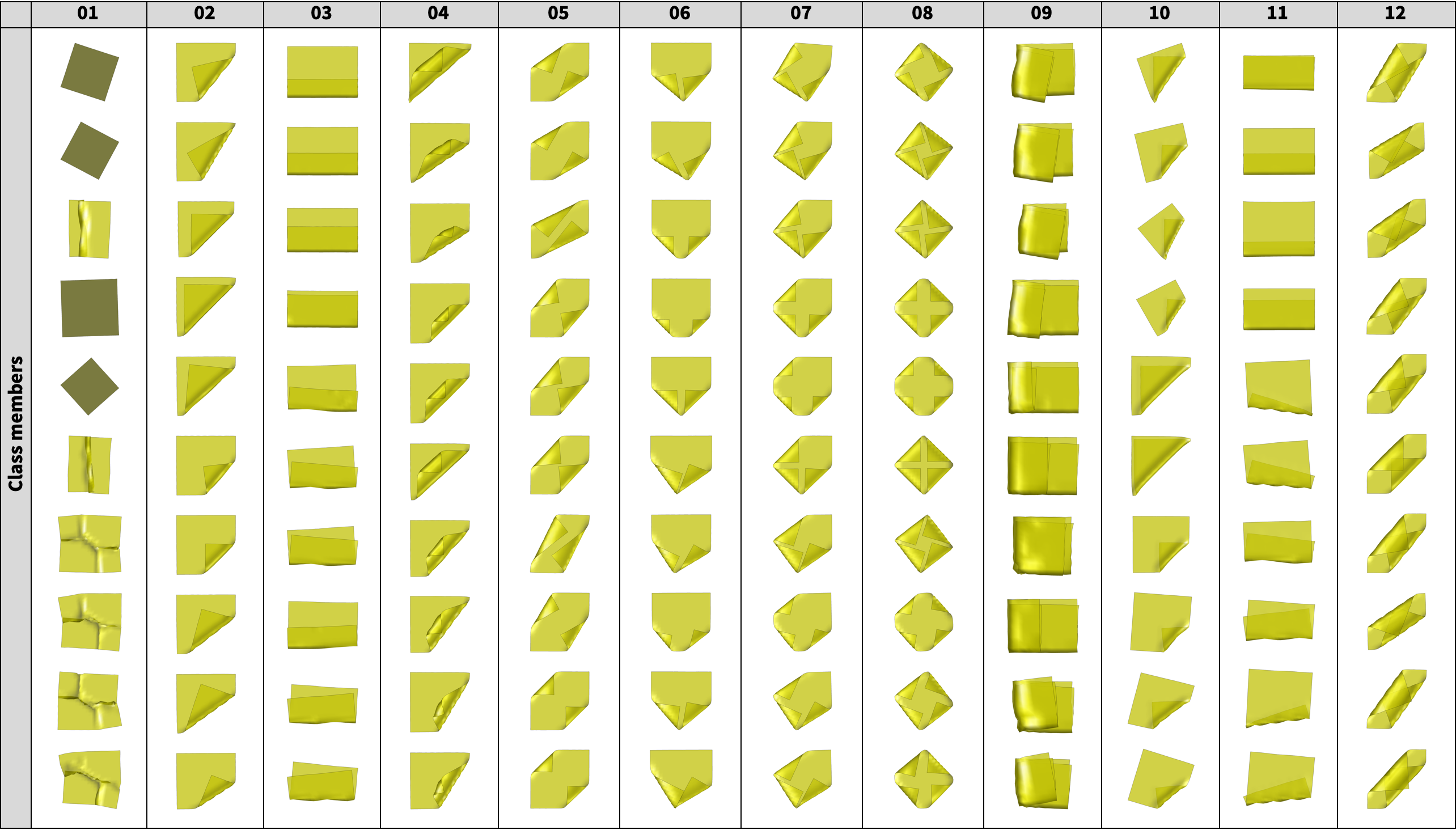}
	
	\vspace{1cm} 
	\includegraphics[width=\linewidth]{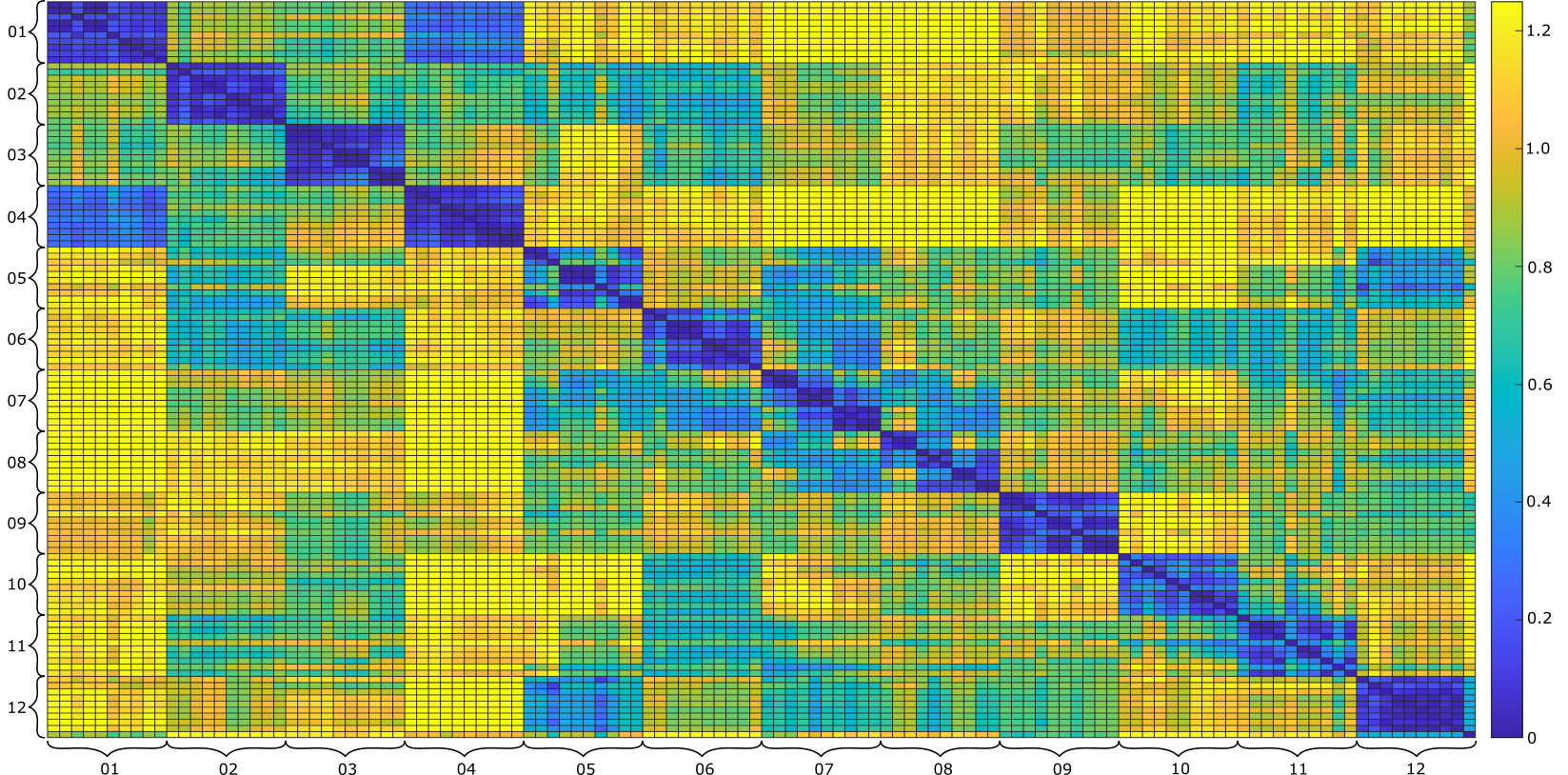}
	\caption{Confusion matrix that computes all the distances between the states shown in the top table. }
	\label{fig:states_and_heatmapGLI}
\end{figure}

In this section we study the ability of the cloth coordinates previously defined to tell apart different cloth states. First, we analyze 3 folding sequences (see Figure \ref{fig:sequences}). We will show that our representation is capable of distinguishing different relevant cloth configurations (e.g. one folded corner vs two folded corners). Then, we will apply our method to a full data-base with 12 cloth classes (shown in Figure \ref{fig:states_and_heatmapGLI}), and we will compare it to 4 alternative representations, proving that ours is more capable to differentiate between cloth states. All data in this section was simulated using the inextensible cloth model described in \cite{COLTRARO2022}.  Finally, we will apply a simple classification method using our representation to real images of folded cloth states.


In order to compare different cloth configurations, once they are represented with our cloth coordinates $\dGLI(\mathcal{C})\in\mathbb{R}^{28}$, it is important to use a proper distance. Due to the scaling factor that we analized in the previous section, the most suitable distance was the Spearman's distance. 
Given two vectors $x,y$ it is defined as
\begin{equation}
 d(x,y) = 
 1-\rho({\operatorname{R}(x),\operatorname{R}(y)}),
\label{eq:SpearmanDistance}
\end{equation}
where $\rho$ is the Pearson correlation coefficient, and $\operatorname{R}(x)$ is the rank variable of $x$. This distance assesses how well the relationship between two vectors $x,y$ can be described using any monotonic function (not only a line). We found this distance to be more sensitive to topological changes than the euclidean distance. This may be due to the fact that this distance focuses on the ranking of bigger coordinates rather than comparing their magnitudes, which is most relevant in our representation. Note that the distance is bounded with values between 0 and 2 and ignores scaling factors between different clothes.

\subsection{Analysis of folding sequences}\label{subsec:sequences}
The first test  compares different cloth states inside a folding sequence. Given the motion of the cloth $\{\mathcal{C}_1,\dots,\mathcal{C}_m\}$, where $m$ is the number of discrete frames and $\mathcal{C}_i$ is the state of the cloth at $t_i$, we compute the confusion matrix $\mathcal{M}_{ij} = d(\dGLI(\mathcal{C}_i),\dGLI(\mathcal{C}_j))$. The 3 folding sequences, shown at the left side of  Figure \ref{fig:sequences} are: folding two opposite corners, folding 4 corners inwards, and folding the cloth in half. The results can be seen on the right side of the figure. Notice how our representation detects changes during the sequence that are topologically meaningful. For example, in Seq. 1, folding two opposite corners, at frame 7, there is a topological change, since a corner changes the orientation from flat to folded, even before its released. This can be seen in the confusion matrix (first two blue squares). This is also clear in Seq. 2 were four corners are folded inwards. Moreover,  our method also detects when edges of the cloth cross (Seq. 1, frame 24, Seq. 3, frame 23). These changes are also meaningful from the manipulation point of view, as they afford different possible graspings or actions.



\subsection{Confusion matrix of the full data-base}

We now analyze a complete data-base consisting of 120 examples classified in 12 different classes of states, shown in Figure \ref{fig:states_and_heatmapGLI}. Most of them are self-explanatory. Note that the class 10 the upper left corner is folded under the cloth (likewise for class 11).
Each class has 10 samples corresponding to the final state of the cloth during a folding sequence simulation.  We manually identified samples that we considered to belong to the same state. We want to emphasize that once we fix an ordering of the corners, our method distinguishes, for example, between different folded corners and this does not contradict the rotational invariance previously shown. 

Again we compute the confusion matrix $\mathcal{M}_{ij} = d(\dGLI(\mathcal{C}_i),\dGLI(\mathcal{C}_j))$ where $\mathcal{C}_k$ is the $k$th example of the data-base. We order the samples, so that the samples from the same classes are consecutive. This way, the plot is more easily interpretable. In Figure \ref{fig:states_and_heatmapGLI} we can see how the classes group without confusion: i.e. the distance between members of a class tends to be smaller (color blue) than the distance to examples outside the class (color yellow).

The confusion matrix shows us interesting insights about our representation. For instance, we can see the two classes 01 and 04 are relatively closer than others. That is because the orientation of all corners is indeed the same in these classes, resulting in a smaller distance in our representation. The same phenomenon can be seen between classes 05 and 12 in some cases, as they are indeed classes with similarities (in 05 the two corners do not cross whereas in 12 they do). However, classes 03 and 11, which differ on whether the folding makes one side of the cloth hide its opposite, are perfectly separated. The borderline cases, that is, the fourth element in class 03 and the first element in class 11 are very similar, but our method distinguishes them because of the relative geometric position between edges (i.e. in these two cases, they are flipped). 
A similar thing occurs between class 02 and 10. It is also worth mentioning that some classes that we have labeled as the same class have clear sub-classes shown in the confusion matrix. That is the case for classes 05, 07 and 08. These are folded corners with different orientations. It is possible using our representation, to induce a partition of the space in order to separate this class into two. In future work we will investigate automatic clustering methods to observe what are the obtained classes, and how we can link different classes under different symmetries that are irrelevant for the affordances.

\begin{table}[htb]
	\caption{Comparison between different representations*}
	\label{tabla_res}
	\begin{tabularx}{\linewidth}{ccccc} \toprule
		\tableheadline{} &  \tableheadline{Database} & \tableheadline{Sequence I} & \tableheadline{Sequence II} & \tableheadline{Sequence III} \\ \midrule
		dGLI      & \textbf{0.73} & \textbf{0.27}  & \textbf{0.18}  & \textbf{0.21} \\ 
		Edges     &         1.60  &         0.68  & 0.77  & 0.51 \\ 
		Corners   &         2.49  &         0.98  & 1.61 &  3.14\\ 
	  Fr\'echet   &         0.99  &         0.69  & 0.76 &  0.48\\ 
		Hausdorff &         1.45  &          0.71  & 0.84 &  0.49\\ 
		\bottomrule
	\end{tabularx}
\vspace{2mm}

*Each number is the Davies-Boulding index introduced in Eq. \ref{eq:DBindex}, that measures cluster separation quality. A smaller value means a better separation. We mark in bold the smallest values in each column.

\end{table}

\subsection{Comparison with other representations} \label{subsec:comparision}

In this section we perform a more quantitative comparison of our representation with other competing methods. To evaluate a representation, we use the standard Davies-Bouldin index to measure cluster separation \cite{indexDB1979}:
\begin{equation}
    DB={\frac {1}{n}}\sum _{i=1}^{n}\max _{j\neq i}\left({\frac {\sigma _{i}+\sigma _{j}}{d(c_{i},c_{j})}}\right)
    \label{eq:DBindex}
\end{equation}
where $n$ is the number of classes (e.g. in the database is 12), $c_{i}$ is the centroid of class $i$ (the average of the coordinates of members of class $i$), $\sigma _{i}$ is a dispersion measure computed as the average distance of all elements in class $i$ to the centroid $c_{i}$ and $d(c_{i},c_{j})$ is the distance between centroids $c_{i}$ and $c_{j}$. With the given classification in Fig. \ref{fig:states_and_heatmapGLI} taken as ground truth, we want a representation that gives a small dispersion inside a class and high distance between the classes, resulting in a low index. The representation and distance with the smallest $DB$ is considered the one that better separates these clusters, and therefore, the best representation to identify different cloth states.

First, we use two simple cloth representations using similar low-level features like the ones we used:
\begin{enumerate}

\item[(i)] \textbf{Edges}: for a given mesh we select the edges shown in Figure \ref{fig:selection_edges} and compute their pairwise minimal distance. This results in a representation vector of lenght 28 just like those of the dGLI coordinates (notice that unlike the dGLI, the coordinates of this vector are always non-negative). We use the Spearman's distance to compare two different samples. This representation is invariant under rigid motions of the plane.

\item[(ii)] \textbf{Corners}: for a given mesh we compute the pairwise distance between its 4 corners. These are 6 non-negative numbers that can be computed for any rectangular cloth, they are invariant by rigid motions and they give a trivial representation of the state of the cloth. We also use Spearman's distance to compare different samples.
\end{enumerate}

 In addition, we compare with two classic methods to measure distance between curves and polygons \cite{alt1995computing,veltkamp2001state}, taking the full discrete boundary curve of the cloth as the representation:
\begin{enumerate}

\item[(iii)] \textbf{Fr\'echet}: to compare two different samples we compute the (discrete) Fr\'echet distance \cite{Frechet1994} between the curves. This is a distance that takes into
account the location and ordering of the points along the curves. Since this distance is not invariant by rigid motions, special care must be taken to center and align the samples before comparing them. In order to do so we center the curves at the origin and perform a rigid alignment by computing the rotation that minimizes the distance between the curves' points.

\item[(iv)] \textbf{Hausdorff}: to compare two different samples we compute the (discrete) Hausdorff distance between the points of the curves \cite{Henrikson1999CompletenessAT}. Informally, two curves are close in the Hausdorff sense if every point of either curve is close to some other point of the other one. This distance disregards the fact that the sets it is comparing are curves and therefore is expected to be less sensitive than the Fr\'echet distance. As before, since this distance is not invariant by rigid motions, we center and align the samples before comparing them. 

\end{enumerate}

In Table \ref{tabla_res} we display the computation of the DB index for our dGLI coordinates and the four discussed methods using as testing scenarios the full database and the 3 folding sequences presented before. As seen in the table our method results in the lowest overall DB in all 4 scenarios, indicating that our method is the one among the studied that best represents the different folded states of the cloths.

\subsection{Real images classification}

Once checked that our method was able to represent folded states of cloth accurately, we implemented a simple classifier of real folded cloth states to highlight its applicability. In order to do so, a representative element of each class in the database shown in Fig. \ref{fig:states_and_heatmapGLI} is chosen, and we estimate the class of a new unclassified sample by choosing its closest representative, using the Spearman distance. 

The real images are taken from a zenital position at 52 cm from the table using a Microsoft Azure Kinect DK 3D camera. A single napkin is used with 3 color stickers attached along each edge close to a corner and on both sides. We first use color segmentation to detect the center of each sticker and get the corresponding 3D point from the depth image. Once all markers are detected, with our combinations of colors on each edge, we can identify each individual corner of the cloth (there are four stickers of the same color around each corner), and its corresponding edge positions, following the same edge selection as in Fig. \ref{fig:selection_edges}.  The obtained size of the observed edges is more than 400 times larger than the edges of the samples of the simulated database, but thanks to the Spearman's distance used, this does not affect the distance values when comparing shapes of different size.

\begin{figure}[h!]
	\centering
	\includegraphics[width=\linewidth]{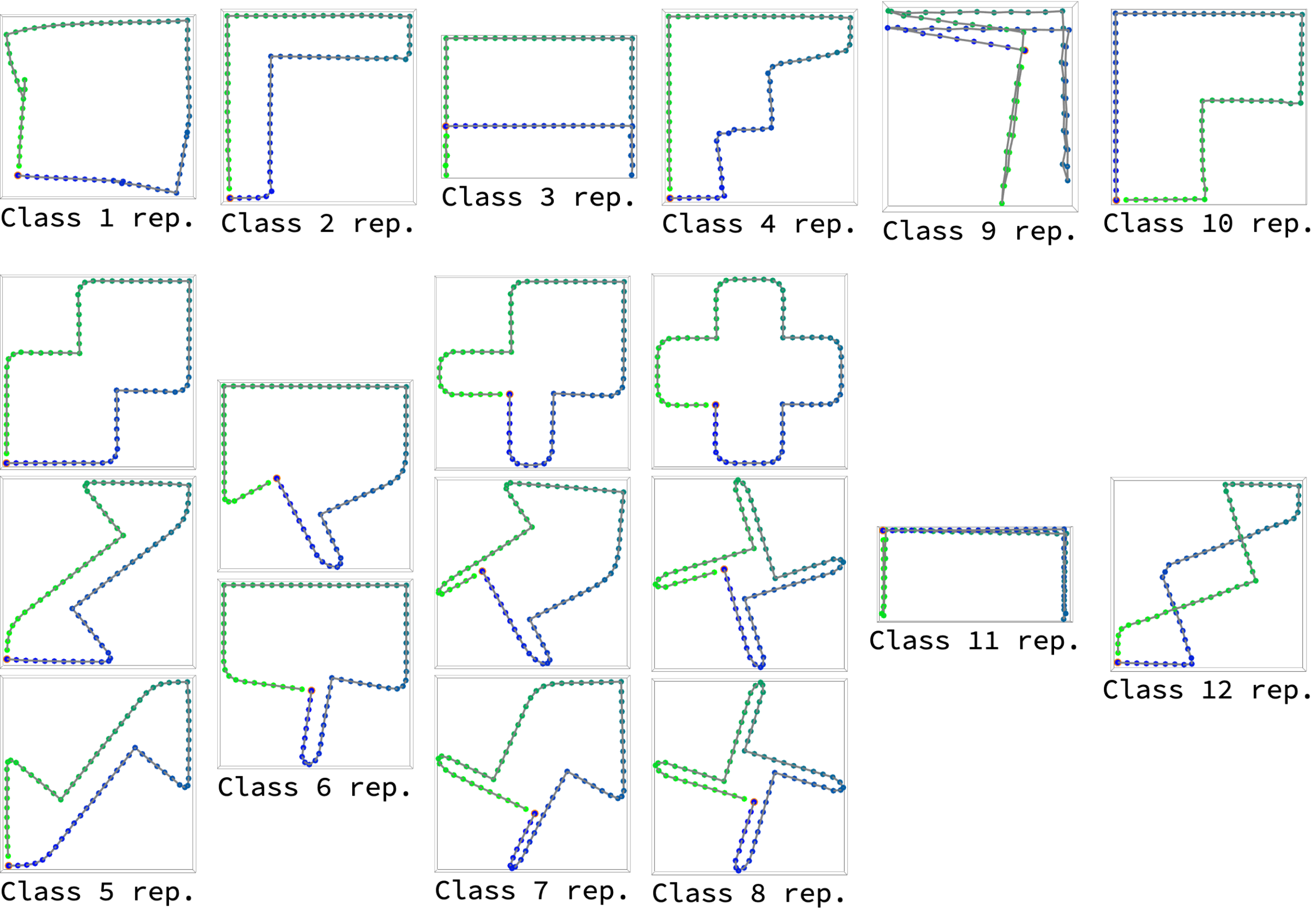}
	\caption{Representatives chosen for each class. When only one is chosen, it is the closest to the centroid of the class. When a class has more sparsity, additional representatives are chosen to represent the subgrups in the class.}
	
	\label{fig:representatives}
\end{figure}

\begin{figure}[h!]
	\centering
	\includegraphics[width=\linewidth]{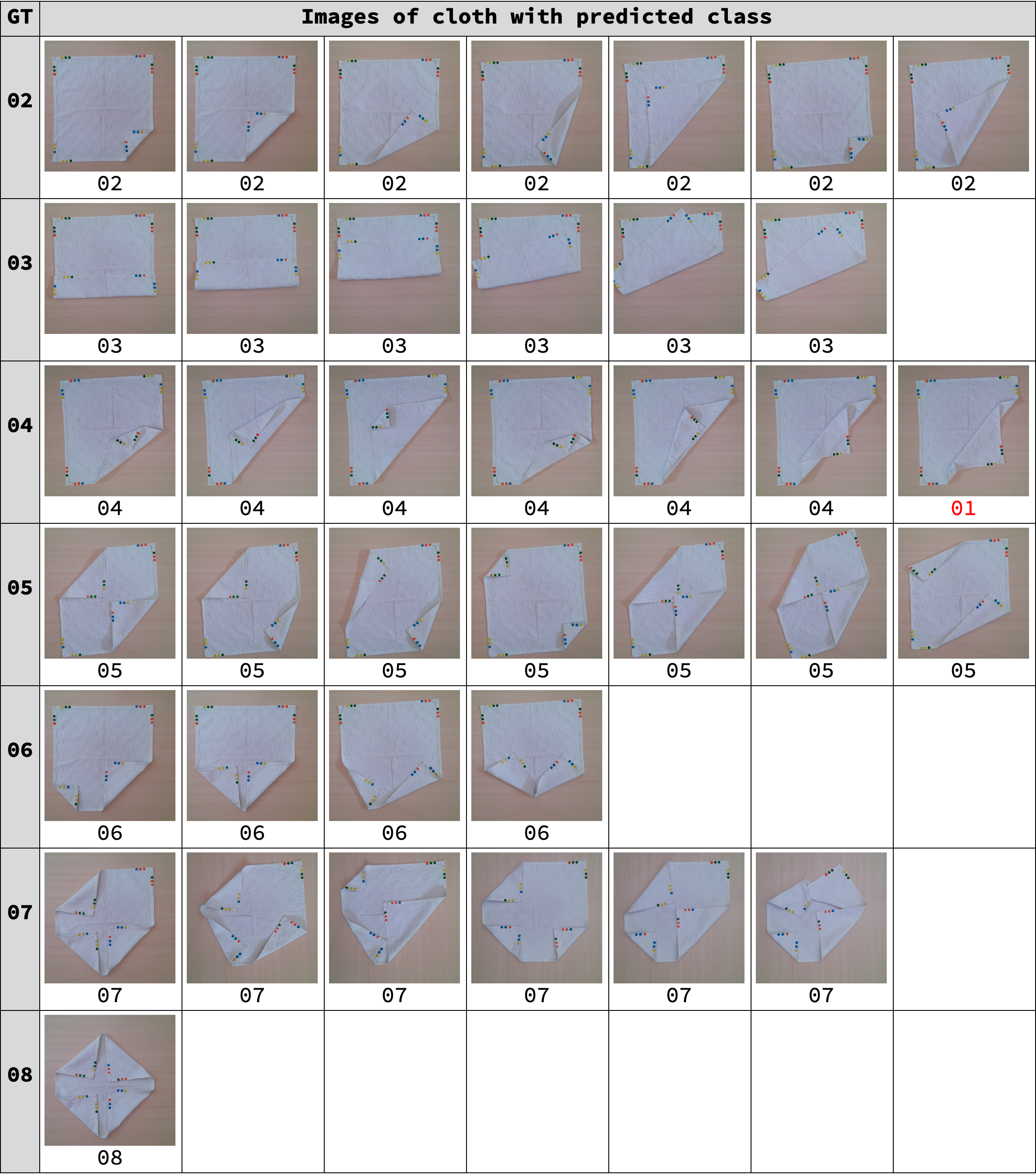}
	\caption{Results of the real image classification using the database presented in Fig. \ref{fig:states_and_heatmapGLI} as reference. The first column shows the ground truth class of the images, and at the bottom of every image the classified class.}
	\label{fig:realImages}
\end{figure}

As we can see in the confusion matrix in Fig. \ref{fig:states_and_heatmapGLI}, some classes have a bigger dispersion in distance because of the variation in orientations of the corners. For these classes, we have chosen 3 different representatives, corresponding to the three different subgroups that can be clearly seen in the confusion matrix. We show the silhouette of the representatives chosen for each class in Fig. \ref{fig:representatives}. The table in Fig. \ref{fig:realImages} shows the results of the classification.

The only miss-classification is the last of the images of the class 04. However, note that is a very extreme case where the cloth is almost flat, and therefore, it is confused with the flat class 01. This is a reasonable mistake, as this cloth can be considered flat enough.

Notice that we can only perceive those textiles with all the stickers visible, therefore, classes with hidden edges, like for instance classes from 09 to 12 where the folding is under the cloth are not present in the real set of samples. However, the classifier still used all 12 classes of the simulated database. This shows that the missing classes don't create confusion in the classification process. In future works, we would like to study the problem of hallucinating the edges of the cloth using deep learning techniques to learn where the hidden edges are, in order to have a bigger and more diverse sample of real images.

\section{Discussion and Conclusions}

We have proposed the dGLI \textit{Cloth Coordinates}, a representation for cloth configurations based on a directional derivative of a topological index that vastly reduces the dimensionality of the configuration space. This reduced representation nevertheless preserves enough information about the configurations to be able to distinguish them based on their grasping affordances using the Spearman distance. The fact that we can identify distinct key configurations in a folding sequence as seen in Figure \ref{fig:sequences} shows great promise for applications in planning for cloth manipulation. Furthermore, our representation allows for different choices of $v$, the perturbation direction, and $\mathcal{S}$, the subset of edges chosen, so that one can fine-tune the representation to the specific task at hand to boost results. Lastly, since our method is not learning based, it does not require any training data, it is completely explainable, and it is robust against possible configurations that are not in the training set. 

Note that the dGLI Cloth Coordinates bridge the gap between low-level features of different cloth configurations, such as the location of corners and edges, to high-level semantic identification of cloth states, associated to the possible affordances. Although a strong assumption is made in this work, that is, that we know the full border of the cloth, there are preliminary works already identifying edges and we are working towards building a data-set to learn to hallucinate such borders when occlusions occur. Meanwhile, our measure can be fully used in simulation with several important applications, such as building data-sets where automatic segmentation of the cloth states is required, monitor cloth manipulation and guide planning methods. We are looking forward to pursue all these lines of research that the present work opens the door to.



Future work also concerns an in-depth analysis of the configuration space defined by our coordinates. In particular we would like to identify a partition of the space that corresponds to a partition of configurations by grasping affordance, which  states are neighbors in this partition, and what the shortest paths from one state to another are. We look forward to carrying out this study analytically as well as through learning methods, which we believe will give better results when the data is enriched and given structure through our representation.

\section{Acknowledgments}
This work was developed in the context of the project CLOTHILDE (``CLOTH manIpulation Learning from DEmonstrations") which has received funding from the European Research Council (ERC) under the European Union's Horizon 2020 research and innovation programme (grant agreement No. 741930).
M. Alberich-Carrami\~nana is also with the Barcelona Graduate School of Mathematics (BGSMath) and the Institut de Matem\`atiques de la UPC-BarcelonaTech (IMTech), and she is partially supported by the grant PID2019-103849GB-I00 funded by MCIN/ AEI /10.13039/501100011033.
 
\bibliographystyle{myplain}
\bibliography{egbibsample}

\end{document}